\pdfoutput=1

\documentclass[11pt]{article}

\usepackage[final]{acl}

\usepackage{times}
\usepackage{latexsym}

\usepackage[T1]{fontenc}

\usepackage[utf8]{inputenc}

\usepackage{microtype}

\usepackage{inconsolata}

\usepackage{graphicx}
\usepackage{booktabs} 
\usepackage{multirow}
\usepackage[arabic,USenglish]{babel}
\newcommand{\ARA}[1]{\begin{footnotesize}\AR{#1}\end{footnotesize}}
\usepackage{adjustbox}

\usepackage{fontawesome5}
\usepackage{listings}

\usepackage[table]{xcolor}  
\usepackage{colortbl}

\usepackage[most]{tcolorbox} 
\usepackage{xcolor}          

\newtcolorbox{promptbox}[1]{
    colback=gray!5,     
    colframe=gray!75,   
    fonttitle=\bfseries,
    title=#1,           
    arc=2mm,            
    outer arc=1mm,
    boxrule=0.5pt,      
    left=10pt,          
    right=10pt,
    top=10pt,
    bottom=10pt
}

\newcommand{\cv}[2]{\cellcolor[gray]{#1}#2}
\usepackage[bb=boondox,bbscaled=.95,cal=boondoxo]{mathalfa}

\newcommand{\modelname}{AraDynFact}

\title{\modelname: Dynamic Evaluation of Factual Knowledge in Arabic}

\author{Ignacio Iacobacci \thanks{corresponding author:\href{mailto:iiacobacci@elm.sa}{iiacobacci@elm.sa}} \\
  Elm Company \\
  \And
  Faroq Altam \\
  Elm Company \\
  \AND
  Zhaozhi Qian \\
  Elm Company \\
  \And
  Muhammad Alqurishi \\
  Elm Company \\
  }

\begin{document}

\maketitle

\begin{abstract}
As Large Language Models (LLMs) continue to scale both in size and capabilities, their proficiency in the Arabic Language has seen significant advancement. However, a critical gap remains: the extent of their factual knowledge and cultural sensitivity to the diverse Arabic-speaking world remains largely underexplored. Current evaluation metrics often focus on translation or generic reasoning, failing to capture the rich historical, social, and regional nuances inherent to Arabic culture.
In addition, most benchmarks rely on heavy work, with human intervention in some steps, making the evaluation of knowledge coverage expensive and slow.
To address this deficiency, we introduce \modelname{}, a novel dynamic evaluation framework designed to rigorously assess the factual Arabic knowledge embedded in LLMs. Unlike static benchmarks, \modelname{} employs a dynamic approach to extract factual information and generate rich and answerable questions in a fast and automatic way.
We apply \modelname{} to Arabic Wikipedia and audit the performance of several state-of-the-art models, ranging from Arabic-centric specialized LLMs to high-resource general purpose LLMs. In addition we found a high degree of correlation with existing, hand-crafted Arabic-centric benchmarks, confirming the potential of our dynamic approach.
\end{abstract}

  

\section{Introduction}

Large language models (LLMs) have achieved remarkable performance in recent years, achieving unprecedented levels of understanding and reasoning in a variety of languages \cite{llama4}. Recent efforts in both general and Arabic-centric LLMs are serving the needs of the Arabic-speaking community.
Numerous benchmarks have been introduced to assess the capabilities and coverage in general knowledge \cite{OALL,alwajih2025palmculturallyinclusivelinguistically,boussaha2025threeLM}, Science \cite[3LM]{boussaha2025threeLM}, Trustworthiness \citep[AraTrust]{alghamdi2025aratrust}, Legal \citep[ALARB]{shairah-etal-2025-alarb}, etc.


These datasets are often static, manually annotated, or derived from existing knowledge in larger LLMs, resulting in the evaluation of effectively stale data (data that is outdated and no longer maintained due to the cessation of data collection or ingestion).

\begin{figure*}[t]
\centering
  \includegraphics[width=1.\textwidth]{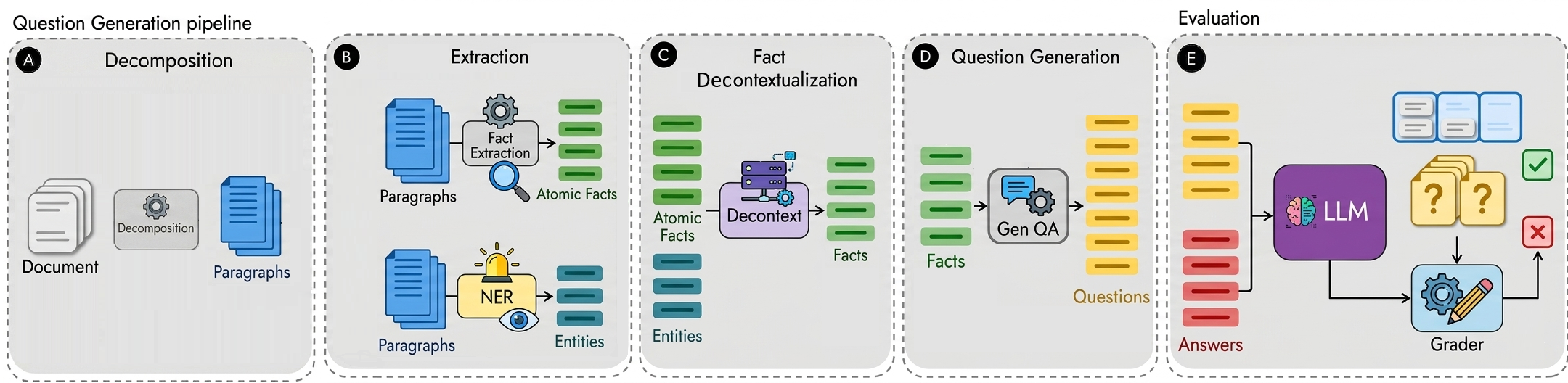}
  \caption{Pipeline of \modelname{}. A) Documents are decomposed into paragraphs. B) From these paragraphs, atomic facts are extracted, and Named Entity Recognition (NER) is performed to identify the most important concepts and entities in the documents. C) Next, fact decontextualization is performed to make the facts self-contained. D) Using both the extracted facts and the original document, questions are then formulated. E) Finally, a grader evaluates the facts, questions, and answers to assess their correctness.}
  \label{fig:pipeline}
\end{figure*}


One potential mitigation for this issue is the emerging trend of dynamic benchmarking \cite{choi-etal-2024-unigen}, which has shown promise as an alternative and has also been proposed to address data leakage.

Some approaches have proposed to construct datasets with a particular cutoff date, \cite{ying2024automating,white2025livebench}, indicating until when the data was effectively collected.
\citet{ying2024automating} proposed a dataset that generates an automatic dataset, while providing online analysis regarding its effectiveness. Others \cite{huang2025datagen} propose to dynamically produce datasets for benchmarking.
These approaches are generally linked to a particular knowledge resource, and most of them are designed to operate exclusively in English.

To address this gap, we propose \modelname{}, an Arabic-centric dynamic benchmark specifically designed to evaluate the factual knowledge of both general-purpose and Arabic-focused language models. We introduce a new pipeline for extracting atomic and relevant facts from a data source, as well as a novel method for formulating questions from one or more of these facts.

Our pipeline has been extensively tested on various Arabic sources, and the resulting benchmark is aligned with the most popular existing Arabic benchmarks. We present an evaluation of several general-purpose and Arabic-centric LLMs and compare their performance with existing Arabic (static) benchmarks.

To our knowledge, no dynamic benchmark has been specifically designed to assess the capabilities of Arabic-capable language models.




Our contributions are threefold:
\begin{itemize}
    \item We introduce \modelname{}, a new benchmark that dynamically generates a set of questions to assess the knowledge coverage of a particular datasource.
    \item We introduce a new pipeline for extracting atomic and relevant facts from a datasource and a new way to formulate questions from one or many of them.
    \item We present an evaluation of several general and Arabic-centric LLMs and compare them with existing Arabic (static) benchmarks.    
\end{itemize}

\section{Related Work}


\subsection{Arabic-specific Evaluation}

The evaluation of Arabic-capable LLMs is an active area of intense research. While many benchmarks are translations of existing ones, some are gathered and designed from scratch. 
Among the most known benchmarks we can cite Open Arabic LLMs Leaderboard \cite[OALL]{OALL} is the de-facto standard evaluation to evaluate the capabilities of LLMs in Arabic. It combines several existing Arabic benchmarks including AlGhafa, ACVA, Arabic MMLU, and Arabic EXAMS, covering tasks such as reading comprehension, sentiment analysis, question answering, and multiple-choice evaluation.
Another recently introduced benchmark is Palm: A Culturally Inclusive and Linguistically Diverse Dataset for Arabic LLMs \citep{alwajih2025palmculturallyinclusivelinguistically}, to date the widest benchmark specifically designed for all varieties of the Arabic Language, covering 20 topics, including culturally informed instructions from dialectal varieties of all 22 Arab countries and from Modern Standard Arabic (MSA). ArabCulture \citep{sadallah2025arabculture} is a culturally grounded commonsense reasoning dataset in Modern Standard Arabic (MSA), covering 13 Arab countries across the Gulf, Levant, North Africa, and the Nile Valley. The dataset contains 3,482 multiple-choice instances that test cultural commonsense reasoning in real-world daily life situations. 
AraTrust \citep{alghamdi2025aratrust} was designed to systematically evaluate how trustworthy large language models are when handling Arabic-language inputs. It focuses on measuring key dimensions of trustworthiness such as factual accuracy, bias, safety, and reliability across different types of Arabic prompts (including cultural and domain-specific questions). 
Finally, 3LM \citep{boussaha2025threeLM} has been introduced, 3LM (\ARA{علم}, science or knowledge in Arabic) is the first Arabic-native benchmark dedicated to scientific reasoning from Arabic educational materials such as biology, physics, chemistry, math, geography and programming.

\subsection{Dynamic Evaluation of LLMs}


\citet{ying2024automating} proposed a method that generates an automatic dataset, thus mitigating the issue of data leakage, while providing online analysis regarding its effectiveness.
DataGen \citep{huang2025datagen} on the other hand, proposes to dynamically produce datasets for benchmarking: a sophisticated framework is designed to produce high-quality synthetic datasets. This approach uses attribute-guided modules and retrieval-augmented techniques to ensure that the information remains factual and diverse. Our approach is similar in spirit to that proposed by \citet{huang2025datagen}, but \modelname{} is specifically aimed at the evaluation of Arabic factual knowledge from potentially any data source.

\subsection{Fact Extraction}

Several approaches include fact extraction, a way to individualize pieces of information that can be verified independently, within their pipelines.
FActScore \citep{min2023factscorefinegrainedatomicevaluation} decomposes long-form generations into a series of atomic facts and computes the percentage of these facts supported by a reliable knowledge resource. The approach estimates factuality by combining retrieval with a strong language model to verify individual claims with high precision. Our Fact extraction process is inspired by their approach.
Another approach that utilizes fact extraction is Factcheck-Bench \citep{wang2024factcheck}. The work introduces a new document-level benchmark to evaluate automatic fact-checkers, focusing on the verifiability of claim-based segments using external evidence retrieved from search engines.

\subsection{Decontextualization}

Decontextualization is the process by which a piece of text can be properly interpreted without the lack of external information. Specifically, to make the piece of text self-contained. Some research pieces have been conducted on this issue from which we drew inspiration while developing \modelname{}.

In the seminal work introduced by \cite{choi-etal-2021-decontextualization}, the authors developed an annotation method and trained automated models to translate regular sentences into a self-contained form. The work output includes a dataset containing triplets (sentence, context, decontextualized\_sentence) that were used to train two different models: i) a BERT-like model to approach the decontextualization as coreference resolution problem, and ii) a SeqToSeq model \cite[T5]{JMLR:v21:20-074}, approaching decontextualization as a translation task.


\section{The \modelname{} Benchmark}

Given a knowledge resource and a model, the objective is to assess how much information from the resource is contained in the model.
The resource is dynamically analyzed to extract factual information. The facts are then used to generate queries that are fed to the model.
Finally, the answers are compared with the original passages from the resource to check their validity. All the steps were carried out using Qwen3-235B-A22B \cite{qwen3technicalreport} with prompts specifically designed to the tasks. The prompts are present in the Appendix.
Figure \ref{fig:pipeline} presents an overview of the whole procedure. Below we will explain in depth each step:



\subsection{Fact Extraction Process}

The fact extraction is itself composed of several sub-tasks: decomposition, atomic fact extraction, named entity recognition, decontextualization and refinement.\\ 
\\
\noindent
\textbf{Decomposition.} The document is divided into paragraphs and later into individual sentences. \\
\textbf{Atomic fact extraction.} From each isolated sentence, the system distills atomic facts—the smallest units of information that can be verified independently.\\
\textbf{Named Entity Recognition.} This process identifies and categorizes salient concepts, such as individuals, organizations, locations, and dates.\\
\textbf{Decontextualization.} From the original paragraph, entities and atomic facts, the latter are rewritten to make them self-contained.\\ 
\textbf{Refinement.} Decontextualized sentences are processed to prevent them from becoming too verbose. \\

\subsection{Question Generation}
\label{sec:qa}

Once the fact list is completely processed we conduct the dynamic generation of questions. 
The process receives a list of facts from a single document and it generates as many self-contained questions as possible. For each question, the process assigns (a) a task taxonomy and (b) a difficulty level.
Four types were used as possible questions tasks:\\

\noindent
\textbf{Factoid QA.} Questions that can be generally answered with an entity. They are generally made from just one fact.\\ 
\textbf{Explanatory QA.} Questions that require an explanation linking two or more facts.\\ 
\textbf{Causal Why/How QA.} The questions from this type are simply formatted as Why/How questions. The answer might not be an entity rather a longer piece of text.\\ 
\textbf{Comparative QA.} The answers of these questions need to address properties of one or more entities.\\ 

\noindent
Each question is also classified in three levels of difficulty, \textit{low}, \textit{medium} and \textit{high}\footnote{In the prompt we used medium, high and 
very\_high as potential options and mapped the three levels accordingly}.

\noindent
\textbf{low}: Direct retrieval or light structuring from selected facts. \\
\textbf{medium}: This option aims to combine multiple facts or requires multi-step reasoning fully supported by the facts.\\
\textbf{high}: Most difficult questions, made with careful constraints, multi-part answer, or subtle synthesis fully determined by the facts.\\

\noindent
Given the strong correlation between the number of facts involved and the reasoning steps required, Factoid questions, typically derived from a single fact, tend to be more straightforward. Although they may still require basic retrieval and understanding, they generally do not demand multi-step reasoning. For this reason, the majority of Factoid questions are classified as \textit{low} difficulty rather than \textit{medium}, as they involve limited compositional reasoning.

\subsection{Answering and grading}
\label{sec:grading}

We follow the standard response generation evaluation strategy. Questions are presented to the models under evaluation using a chat template.
Since by construction the questions are self-contained, no further context is provided. The knowledge needed to answer the questions should be encoded already within model's weights. In this way, given a data source, we can dynamically assess the knowledge coverage of any language model.

\subsubsection*{LLM-as-a-Judge}

Once all the questions have been answered by the models under evaluation, the grading process begins. Because the questions are dynamically generated, there is no single canonical “gold” answer for each one. Instead, the evaluation relies on the set of facts used to generate the question, which serves as the reference for assessing the correctness of the responses. We also rely on Qwen3-235B-A22B \cite{qwen3technicalreport} as our Judge model. For evaluation itself, we adopt the strategy, and used the prompt, introduced in \citet[SimpleQA]{wei2024measuringshortformfactualitylarge}, where each model-generated answer is assigned one of three labels: \textbf{CORRECT}, \textbf{INCORRECT}, or \textbf{NOT\_ATTEMPTED}.  
In the case of \textbf{CORRECT} responses, the answer must fully incorporate all the key information from the fact list used to create the question, while maintaining internal consistency. \textbf{INCORRECT} responses are those that either provide partial information, omit critical facts, or contain statements that directly contradict the reference facts. Finally, responses labeled \textbf{NOT\_ATTEMPTED} correspond to cases where the model explicitly indicates uncertainty (e.g., by stating “I don’t know”) or Produces a response that introduces no factual content beyond what is already contained in the question.

\begin{table*}[t]
  \centering
  \begin{adjustbox}{width=\textwidth,center}
   \normalsize
   \renewcommand{\arraystretch}{1.05}
  \begin{tabular}{l| l| ccc|ccc|ccc|ccc }
    \toprule
    \multirow{3}{*}{Lang} & \multirow{3}{*}{Model}  & \multicolumn{3}{c|}{Factoid QA} & \multicolumn{3}{c|}{Explanatory QA}	&\multicolumn{3}{c|}{Causal Why/How QA} & \multicolumn{3}{c}{Comparative QA} \\
    
&	&	low	&	medium	&	high	&	low	&	medium	&	high	&	low	&	medium	&	high	&	low	&	medium	&	high \\
&	&	4263 	&	384	&	6	&	377	&	1072	&	90	&	409	&	4619	&	460	&	331	&	986	&	67
 \\
\midrule

\multirow{5}{*}{\rotatebox[origin=c]{90}{Arabic}}
  &Fanar-1-9B
    & \cv{0.70}{0.34} & \cv{0.69}{0.35} & \cv{0.85}{0.17}
    & \cv{0.72}{0.31} & \cv{0.70}{0.34} & \cv{0.64}{0.40}
    & \cv{0.70}{0.34} & \cv{0.70}{0.34} & \cv{0.67}{0.37}
    & \cv{0.71}{0.32} & \cv{0.69}{0.35} & \cv{0.68}{0.36} \\
  &ALLaM-7B-Instruct-preview
    & \cv{0.72}{0.31} & \cv{0.70}{0.33} & \cv{0.40}{0.67}
    & \cv{0.70}{0.33} & \cv{0.70}{0.34} & \cv{0.70}{0.34}
    & \cv{0.72}{0.31} & \cv{0.70}{0.33} & \cv{0.72}{0.31}
    & \cv{0.77}{0.26} & \cv{0.69}{0.35} & \cv{0.70}{0.34} \\
  &Yehia-7B-preview
    & \cv{0.72}{0.31} & \cv{0.70}{0.33} & \cv{0.70}{0.33}
    & \cv{0.72}{0.31} & \cv{0.70}{0.33} & \cv{0.66}{0.38}
    & \cv{0.71}{0.32} & \cv{0.70}{0.33} & \cv{0.73}{0.30}
    & \cv{0.74}{0.29} & \cv{0.69}{0.35} & \cv{0.70}{0.34} \\
  &SILMA-9B-Instruct-v1.0
    & \cv{0.78}{0.25} & \cv{0.75}{0.28} & \cv{0.85}{0.17}
    & \cv{0.78}{0.25} & \cv{0.75}{0.28} & \cv{0.76}{0.27}
    & \cv{0.77}{0.26} & \cv{0.77}{0.26} & \cv{0.79}{0.23}
    & \cv{0.75}{0.28} & \cv{0.77}{0.26} & \cv{0.75}{0.28} \\
  &Command R7B Arabic
    & \cv{0.69}{0.35} & \cv{0.72}{0.31} & \cv{1.00}{0.00}
    & \cv{0.74}{0.29} & \cv{0.69}{0.35} & \cv{0.65}{0.39}
    & \cv{0.69}{0.35} & \cv{0.69}{0.35} & \cv{0.69}{0.35}
    & \cv{0.71}{0.32} & \cv{0.65}{0.39} & \cv{0.65}{0.39} \\
\midrule
\multirow{12}{*}{\rotatebox[origin=c]{90}{General}}
  &Qwen3 8B
    & \cv{0.73}{0.30} & \cv{0.71}{0.32} & \cv{0.85}{0.17}
    & \cv{0.73}{0.30} & \cv{0.75}{0.28} & \cv{0.72}{0.31}
    & \cv{0.78}{0.25} & \cv{0.74}{0.29} & \cv{0.73}{0.30}
    & \cv{0.76}{0.27} & \cv{0.73}{0.30} & \cv{0.75}{0.28} \\
  &Qwen3 32B
    & \cv{0.72}{0.31} & \cv{0.73}{0.30} & \cv{0.70}{0.33}
    & \cv{0.74}{0.29} & \cv{0.71}{0.32} & \cv{0.80}{0.22}
    & \cv{0.73}{0.30} & \cv{0.71}{0.32} & \cv{0.69}{0.35}
    & \cv{0.71}{0.32} & \cv{0.73}{0.30} & \cv{0.70}{0.33} \\
  &Qwen3 30B A3B
    & \cv{0.72}{0.31} & \cv{0.71}{0.32} & \cv{0.85}{0.17}
    & \cv{0.72}{0.31} & \cv{0.72}{0.31} & \cv{0.68}{0.36}
    & \cv{0.73}{0.30} & \cv{0.72}{0.31} & \cv{0.72}{0.31}
    & \cv{0.74}{0.29} & \cv{0.72}{0.31} & \cv{0.70}{0.33} \\
  &Llama-3.1-8B-Instruct
    & \cv{0.82}{0.20} & \cv{0.83}{0.19} & \cv{0.55}{0.50}
    & \cv{0.83}{0.19} & \cv{0.82}{0.20} & \cv{0.86}{0.16}
    & \cv{0.83}{0.19} & \cv{0.81}{0.21} & \cv{0.82}{0.20}
    & \cv{0.86}{0.16} & \cv{0.82}{0.20} & \cv{0.83}{0.19} \\
  &Llama-3.3-70B-Instruct
    & \cv{0.65}{0.39} & \cv{0.67}{0.37} & \cv{0.70}{0.33}
    & \cv{0.68}{0.36} & \cv{0.63}{0.41} & \cv{0.61}{0.43}
    & \cv{0.64}{0.40} & \cv{0.64}{0.40} & \cv{0.62}{0.42}
    & \cv{0.61}{0.43} & \cv{0.65}{0.39} & \cv{0.62}{0.42} \\
  &Llama-4-Scout-17B-16E
    & \cv{0.68}{0.36} & \cv{0.66}{0.38} & \cv{1.00}{0.00}
    & \cv{0.66}{0.38} & \cv{0.67}{0.37} & \cv{0.64}{0.40}
    & \cv{0.69}{0.35} & \cv{0.68}{0.36} & \cv{0.71}{0.32}
    & \cv{0.67}{0.37} & \cv{0.68}{0.36} & \cv{0.65}{0.39} \\
  &DeepSeek-R1-Distill-Llama-8B
    & \cv{0.84}{0.18} & \cv{0.84}{0.18} & \cv{0.85}{0.17}
    & \cv{0.83}{0.19} & \cv{0.83}{0.19} & \cv{0.86}{0.16}
    & \cv{0.80}{0.22} & \cv{0.83}{0.19} & \cv{0.83}{0.19}
    & \cv{0.85}{0.17} & \cv{0.82}{0.20} & \cv{0.80}{0.22} \\
  &DeepSeek-R1-Distill-Llama-70B
    & \cv{0.64}{0.40} & \cv{0.65}{0.39} & \cv{0.85}{0.17}
    & \cv{0.66}{0.38} & \cv{0.63}{0.41} & \cv{0.64}{0.40}
    & \cv{0.65}{0.39} & \cv{0.63}{0.41} & \cv{0.61}{0.43}
    & \cv{0.62}{0.42} & \cv{0.66}{0.38} & \cv{0.57}{0.48} \\
  &gemma-3-12b-it
    & \cv{0.68}{0.36} & \cv{0.69}{0.35} & \cv{1.00}{0.00}
    & \cv{0.69}{0.35} & \cv{0.66}{0.38} & \cv{0.62}{0.42}
    & \cv{0.67}{0.37} & \cv{0.66}{0.38} & \cv{0.70}{0.34}
    & \cv{0.69}{0.35} & \cv{0.64}{0.40} & \cv{0.64}{0.40} \\
  &gemma-3-27b-it
    & \cv{0.64}{0.40} & \cv{0.68}{0.36} & \cv{0.70}{0.33}
    & \cv{0.67}{0.37} & \cv{0.63}{0.41} & \cv{0.64}{0.40}
    & \cv{0.64}{0.40} & \cv{0.63}{0.41} & \cv{0.62}{0.42}
    & \cv{0.63}{0.41} & \cv{0.63}{0.41} & \cv{0.67}{0.37} \\
  &gpt-oss-20b
    & \cv{0.68}{0.36} & \cv{0.67}{0.37} & \cv{1.00}{0.00}
    & \cv{0.72}{0.31} & \cv{0.70}{0.33} & \cv{0.68}{0.36}
    & \cv{0.69}{0.35} & \cv{0.68}{0.36} & \cv{0.68}{0.36}
    & \cv{0.66}{0.38} & \cv{0.69}{0.35} & \cv{0.68}{0.36} \\
  &gpt-oss-120b
    & \cv{0.62}{0.42} & \cv{0.61}{0.44} & \cv{0.85}{0.17}
    & \cv{0.67}{0.37} & \cv{0.63}{0.41} & \cv{0.63}{0.41}
    & \cv{0.60}{0.45} & \cv{0.62}{0.42} & \cv{0.61}{0.44}
    & \cv{0.62}{0.42} & \cv{0.62}{0.42} & \cv{0.73}{0.30} \\

    \bottomrule
  \end{tabular}

  \end{adjustbox}
    \caption{Accuracy of the experiments carried out on Arabic Wikipedia with thefour types of Questions varying question difficulty. High-difficulty Factoid questions are rare (only 6 occurrences).}
  \label{tab:stats}
\end{table*}

\section{Experiments}

We tested our pipeline on 2026 dump of Arabic Wikipedia. The Arabic Wikipedia\footnote{https://en.wikipedia.org/wiki/Arabic\_Wikipedia.} (\ARA{ويكيبيديا العربية}), is the version of Wikipedia written in Modern Standard Arabic. As of March 2026, it contained more than 1.3 million articles, ranking 15th in terms of number of articles among Wikipedias. As versions from different languages, articles are generally attached with categories. We kept all articles linked at first or second level from the category Middle East (\ARA{تصنيف:الشرق الأوسط}) resulting in almost 60k articles.
These articles contained approximately 600k paragraphs, where 10 million atomic facts were extracted.

The question generation process produced approximately 340k different questions and we sampled uniformly 5\% of the corpus and filtered trivial (extremely short) questions, resulting in 13064 questions, which we used in our experiments.

\subsection{Results}

Table \ref{tab:stats} shows an analysis of the four types of questions run on 17 standard models, both Arabic-centric and general models.
For Arabic models, we analyze Fanar \cite{fanarllm2025}, ALLaM \cite{bari2025allam}, Yehia \cite{yehia2025}, SILMA \cite{silma-9b-2024} and Command-R7B-Arabic \cite{alnumay2025command}. For the English models, we included different sizes of Qwen3 \cite{qwen3technicalreport}, Llama \cite{grattafiori2024llama3herdmodels,llama4}, two distilled versions of DeepSeek R1 \cite{guo2025deepseek}, Gemma 3 \cite{gemma_2025} and GPT-OSS \cite{agarwal2025gpt}

Table \ref{tab:stats} presents the performance of 17 large language models across the four question types defined in \modelname{} (Factoid QA, Explanatory QA, Causal Why/How QA, and Comparative QA), each broken down by three difficulty levels: low, medium and high. Below the difficulty, we include the number of questions generated for each type/difficulty. As it is easy to notice, Factoid questions tend to be easier than the other types, which is represented in the bias towards low-difficulty questions. The majority of the questions from other types fall into the medium difficultly. The rarity of high-difficulty Factoid QA questions makes the indicator for that category unreliable. The remaining categories show a more reasonable distribution.

Regarding the models, they are grouped into two categories: Arabic-centric (Fanar-1-9B, ALLaM-7B, Yehia-7B, SILMA-9B, and Command-R7B-Arabic) and general-purpose (various sizes of Qwen3, Llama, DeepSeek-R1 distillations, Gemma 3, and GPT-OSS).
Overall scores remain modest in all models and question types, with most values falling in the 0.20 to 0.45 range, reflecting the inherent difficulty of the benchmark. Among the top performers, larger general models such as GPT-OSS-120B, DeepSeek-R1-Distill-Llama-70B, Llama-3.3-70B, and Gemma-3-27B consistently achieve the highest scores, suggesting that scale remains a dominant factor even for Arabic factual knowledge. In contrast, smaller models, both Arabic-centric and general, such as SILMA-9B, Llama-3.1-8B, and DeepSeek-R1-Distill-Llama-8B, tend to underperform across all question types.
Arabic-centric models generally outperform general-purpose models of comparable size, with the exception of SILMA-9B. This indicates that, while specialized Arabic training may be beneficial, it does not necessarily guarantee superior factual coverage on this benchmark.
Performance on high-difficulty questions is particularly volatile, with several models scoring near zero, likely due to the limited number of instances at that level. Scores are broadly consistent across question types within each model, though Comparative and Causal questions tend to surface slightly more variance, consistent with their higher compositional reasoning demands.

\section{Analysis}

\subsection*{Does \modelname{} correlate with models’ strength?}

\begin{figure}[t]
  \centering
  \includegraphics[width=0.85\columnwidth]{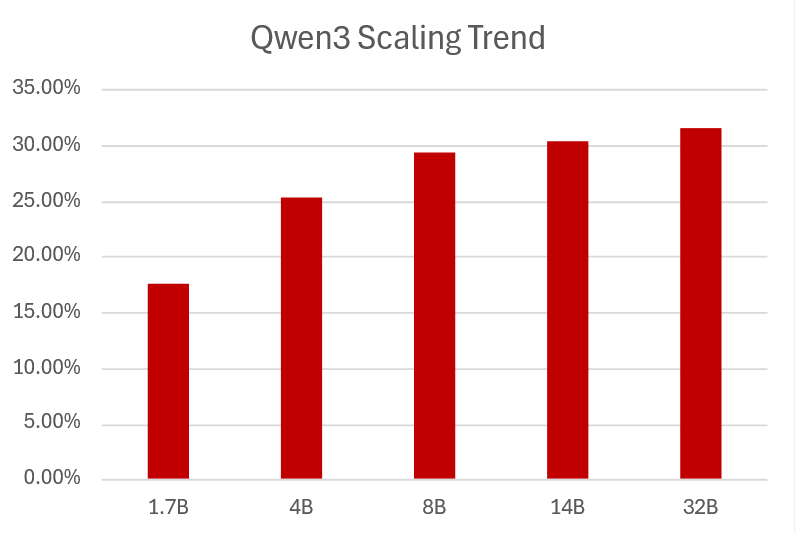}
  \caption{Visualization Qwen3 performance on increasingly model sizes.}
  \label{fig:qwen}
\end{figure}

To validate this premise, we conducted two complementary experiments designed to assess both the internal consistency of our benchmark and its external alignment with established evaluation frameworks.
The first experiment focuses on scaling behavior. Specifically, we selected a model family released across multiple parameter sizes while sharing the same architecture, training data, and optimization regime. This setup allows us to control for confounding variables, since model size remains the primary factor that changes across variants. Under standard scaling laws, increasing the number of parameters should generally lead to improved performance, provided the evaluation benchmark is sufficiently sensitive to capture differences in reasoning capacity. Therefore, if \modelname{} is a reliable and well-calibrated benchmark, it should reflect a consistent and monotonic performance improvement as model size increases. In other words, larger models should systematically outperform their smaller counterparts. As illustrated in Figure \ref{fig:qwen}, the results confirm this expectation: performance on \modelname{} improves steadily with model scale, demonstrating that the benchmark is sensitive to model scale and consistent with established scaling laws.

\subsection*{Does \modelname{} correlate with existing benchmarks?}

Our second verification step evaluates external validity by examining how \modelname{} correlates with existing Arabic-focused benchmarks. To this end, we selected a representative subset of tasks from the Open Arabic LLM Leaderboard \cite[OALL]{OALL}, including ArabicMMLU, Exams, MedinahQA, and AraTrust. These benchmarks collectively cover general knowledge, academic-style examinations, question answering, and trustworthiness evaluation. In addition, we incorporated two previously discussed benchmarks: 3LM \cite{boussaha2025threeLM}, which emphasizes STEM-oriented reasoning, and ArabCulture \cite{sadallah2025arabculture}, which focuses on culturally grounded commonsense reasoning.

\begin{figure}[t]
  \centering
  \includegraphics[width=0.85\columnwidth]{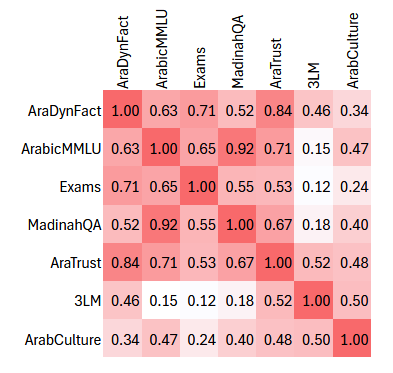}
  \caption{Spearman correlation between different bench-
mark suites}
  \label{fig:spearman}
\end{figure}

We computed the Spearman rank correlation coefficient across multiple large language models. As shown in Figure \ref{fig:spearman}, the results indicate strong positive correlations, suggesting that \modelname{} is well aligned with recognized evaluation standards. Importantly, unlike many static benchmarks, \modelname{} offers the additional advantage of dynamic generation, enabling continuous expansion and reduced risk of data leakage while maintaining agreement with established evaluation signals.



\section{Conclusion}

In this work, we introduced \modelname{}, a dynamic evaluation framework for assessing the factual Arabic knowledge embedded within LLMs. By grounding evaluation in automatically extracted atomic facts from potentially any Arabic data source, and generating diverse questions across four types and three difficulty levels, \modelname{} offers a scalable and contamination-resistant alternative to static benchmarks.
Our comprehensive evaluation using Arabic Wikipedia as knowledge resource, across 17 models reveals a consistent pattern: while general-purpose LLMs often achieve competitive scores on linguistic and reasoning tasks, they exhibit notable gaps in localized factual accuracy and culturally specific knowledge about the Arabic-speaking world. Arabic-centric models, despite being trained on domain-relevant data, do not uniformly outperform their general counterparts, highlighting that scale and general pretraining remain strong factors even in culturally specific evaluation settings.

We demonstrated that \modelname{} performance scales monotonically with model size within the same model family, confirming the benchmark's sensitivity and reliability. Second, strong Spearman rank correlations with established Arabic benchmarks such as ArabicMMLU, AraTrust, and ArabCulture confirm that \modelname{} aligns well with recognized evaluation signals, while offering the added benefit of dynamic regeneration.

We hope \modelname{} serves as a foundation for building more culturally aware and factually reliable Arabic AI systems, and encourage the community to extend the framework to additional Arabic knowledge sources and dialects beyond Modern Standard Arabic.

\newpage

\section*{Limitations}

While our method shows potential and does correlate with existing datasets, there are limitations that are intrinsic to the pipeline process. Questions are generated from facts, answered by evaluated models and graded by a larger LLM. There are some occasions where the question is not completely answerable, or the grader fails to correctly judge the validity of an answer or identify its errors. We consider that in the long run, those errors even out and a reasonable evaluation of the factual knowledge is performed. 




\bibliography{custom}

\appendix

\section{Example}

\begin{figure*}[t]
\centering
  \includegraphics[width=1.\textwidth]{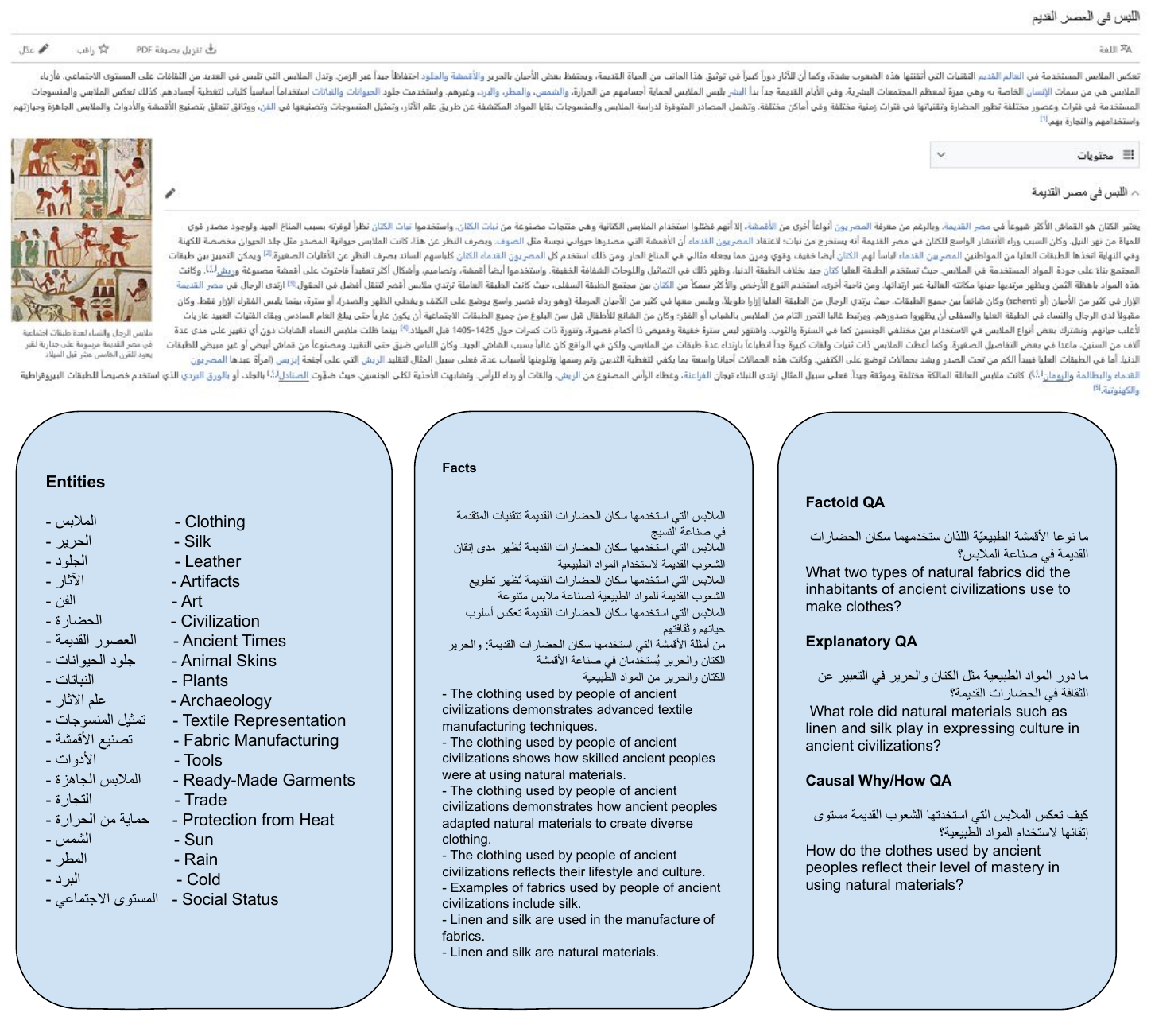}
  \caption{Excerpt of how \modelname{} process an example document, the Arabic page of \href{https://ar.wikipedia.org/wiki/\%D8\%A7\%D9\%84\%D9\%84\%D8\%A8\%D8\%B3\_\%D9\%81\%D9\%8A\_\%D8\%A7\%D9\%84\%D8\%B9\%D8\%B5\%D8\%B1\_\%D8\%A7\%D9\%84\%D9\%82\%D8\%AF\%D9\%8A\%D9\%85}{Clothing in the ancient world}", to produce questions}
  \label{fig:excerpt}
\end{figure*}

\section{Prompts used}
\begin{figure*}
\begin{promptbox}{Named Entity Extraction}

Please list the *key* entities are referred in this text. \\
List only up to 20 most important entities. Use the same language used in the text:  [PARAGRAPH] \\
\\
=== OUTPUT FORMAT===\\
- A list starting with the '-' symbol.\\
- Disregard any punctuation.\\
\end{promptbox}
\end{figure*}

\begin{figure*}
\begin{promptbox}{Atomic Fact Extraction}
Please breakdown the following sentence into independent facts in the same language of the original sentence: [EXAMPLE FACTS] \\

=== Question ===\\
Please breakdown the following sentence into independent facts same language of the original sentence: [SENTENCE]\\

=== Output Format ===\\
The list must follow the same style as the examples. \\
The facts should be written in Arabic. \\
You MUST use "-" to numerate each extracted fact. \\
Do not use numbers or any other symbols. \\
=== Final Answer ===\\
\end{promptbox}
\end{figure*}

\begin{figure*}

\begin{promptbox}{Decontextualizer}
Using this this context: [PARAGRAPH]\\
\\
and this list of entities: [ENTITIES]\\
\\
Please rewrite this text snippet *as a single sentence* by adding references to the entities the snippet might be referring to. Do not add extra facts to the snippet: [FACTS] \\
\\
=== OUTPUT FORMAT===\\
- Do NOT add extra facts to those contained in the snippet.\\
- A **single sentence** without any missing references.\\
- Any pronoun should have a reference within the sentence.\\
- All entities should be named or clearly disambiguated.\\
- Avoid any reference to the text. No sentences such as 'in this text' or 'according to the text'. \\
\\
=== Final Answer ===\\
\end{promptbox}
\end{figure*}

\begin{figure*}
\small
\begin{promptbox}{Question Generation}

------------------------------------------------------------------------------------------------------------- \\
SYSTEM: Facts List → Self-Contained Questions (YAML only) \\
-------------------------------------------------------------------------------------------------------------\\
\\
You receive FACTS as a plain list of strings. Your job is to generate MANY self-contained questions that are answerable ONLY from the selected facts, and output ONE YAML document only.\\
\\
-------------------------------------------------------------------------------------------------------------\\
HARD REQUIREMENTS (must always hold)\\
-------------------------------------------------------------------------------------------------------------\\
\\
R1) Output must be ONE valid YAML document only. \\
     - No markdown fences, no explanations, no extra text. \\
     - Output YAML starting immediately with the first key (e.g., "questions:"), and nothing before it. \\
\\
R2) Absolute prohibition on referencing the input or "facts" in the question text: \\
     - The question\_text MUST NOT contain ANY wording that points to external context or provided material, including (but not limited to): \\
       * "\ARA{اعتمادًا على}", "\ARA{استنادًا إلى}", "\ARA{حسب}", "\ARA{وفق}", "\ARA{بناءً على/بناءا على}" \\
       * "\ARA{كما جاء}", "\ARA{كما ورد}", "\ARA{المذكور أعلاه}", "\ARA{المعلومات المذكورة}", "\ARA{المعلومات المعطاة/المعطاه}" \\
       * "\ARA{البيانات}", "\ARA{المصدر}", "\ARA{المدخل}", "\ARA{النص}", "\ARA{القائمة}", "\ARA{الحقائق}", "ID", "\ARA{المرفق}" \\
       * English equivalents: "based on", "according to", "given", "provided", "above", "the facts", "the list", "the text", "input", "source", "context" \\
     - If a draft question includes any such phrase, you MUST rewrite it to be fully self-contained or discard it.
\\
R3) Every question MUST use at least one fact from FACTS. \\
     - Each question must be grounded in the content of its selected fact(s). \\
\\
R4) Every question MUST be answerable ONLY from the selected facts used to build it: \\
     - Do not ask for anything that requires outside knowledge, missing context, assumptions, opinions, or web searching. \\
     - If a question would normally need external knowledge, do NOT generate it. \\
\\
R5) Questions must be INCLUSIVE of the facts: \\
    - Prefer questions that combine multiple facts when the answer remains strictly determined by those facts. \\
    - Do not invent details. Do not contradict any fact. \\
\\
R6) Generate as many questions as possible: \\
    - Produce the maximum number of distinct, non-duplicate questions you can, given FACTS. \\ 
    - Each question must still satisfy R2–R5. \\
\\
R7) No hallucinations: \\
    - Do not add names, dates, numbers, locations, causes, or outcomes not explicitly present in the selected facts. \\
    - Only allow "inferred\_pattern" when it is a strict logical consequence of the selected facts. \\
\\
-------------------------------------------------------------------------------------------------------------\\
INPUT FORMAT \\
-------------------------------------------------------------------------------------------------------------\\
\\
FACTS is provided directly here: \\
\\
{facts} \\
\\
-------------------------------------------------------------------------------------------------------------\\
TASK TAXONOMY (pick ONE per question) \\
-------------------------------------------------------------------------------------------------------------\\
\\
Use ONE of these (exactly): \\
\\
1 Factoid QA \\
2 Explanatory QA \\
3 Comparative QA \\
4 Causal Why/How QA \\
\\
Store: \\
- task\_taxonomy\_id (e.g., "4") \\
- task\_taxonomy\_name (exact name from list above) \\
\\
\end{promptbox}
\end{figure*}

\begin{figure*}
\small
\begin{promptbox}{Question Generation (cont..)}

-------------------------------------------------------------------------------------------------------------\\
DIFFICULTY (pick ONE per question) \\
-------------------------------------------------------------------------------------------------------------\\
\\
difficulty: "medium" | "high" | "very\_high" \\
\\
- medium: direct retrieval / light structuring from selected facts \\
- high: combines multiple facts or requires multi-step reasoning fully supported by the facts \\
- very\_high: careful constraints, multi-part answer, or subtle synthesis still fully determined by the facts \\
\\
-------------------------------------------------------------------------------------------------------------\\
OUTPUT YAML SCHEMA (simplified) \\
-------------------------------------------------------------------------------------------------------------\\
\\
Return exactly this YAML structure: \\
\\
questions: \\
   - id: 1\\
    question\_text: "..." \\
    source\_facts\_ids: [1, 2] \\
    task\_taxonomy\_id: "4"\\
    task\_taxonomy\_name: "Causal Why/How QA"\\
    difficulty: "high"\\
\\
meta: \\
    facts\_count: 0\\
    generated\_questions\_count: 0\\ 
    task\_taxonomy\_ids\_summary: ["1", "4"]\\
    difficulty\_summary: ["medium", "high", "very\_high"]\\
\\
checklist:\\
    yaml\_only: true\\
    questions\_have\_no\_input\_references: true\\
    each\_question\_self\_contained: true\\
    each\_question\_uses\_at\_least\_one\_fact: true\\
    each\_question\_answerable\_from\_selected\_facts\_only: true\\
    no\_external\_knowledge\_required: true\\
    no\_duplicates: true\\
\\
-------------------------------------------------------------------------------------------------------------\\
INTERNAL PROCESS (do not print)\\
-------------------------------------------------------------------------------------------------------------\\
\\
1) Index FACTS from 1..N.\\
2) Generate the maximum number of distinct questions that satisfy the hard requirements.\\
3) For each question, choose the minimal set of facts that fully determines the answer.\\
4) Write question\_text with ZERO references to any input/provided material (see R2 prohibited phrases).\\
5) Assign taxonomy and difficulty that match the question.\\
6) Populate meta fields and checklist booleans.\\
7) Final self-check:\\
   - YAML only\\
   - question\_text contains none of the prohibited reference phrases in ANY language\\
   - Each question uses >=1 fact\\
   - Each question answerable only from its selected facts\\
   - No duplicates\\
   - If any checklist item would be false, fix questions until all are true.\\
\\
-------------------------------------------------------------------------------------------------------------\\
NOW DO THE TASK\\
-------------------------------------------------------------------------------------------------------------\\
\\
Use the provided FACTS list and output the final YAML document only.\\
\\
\end{promptbox}
\end{figure*}

\end{document}